\documentclass[10pt,a4paper,twocolumn]{article}

\usepackage{hyperref}
\usepackage{graphicx}
\usepackage{booktabs}
\usepackage{tabularx}
\usepackage{authblk}
\usepackage[backend=bibtex,style=numeric,sorting=nyt]{biblatex}
\begin{document}

\title{Automated title and abstract screening for scoping reviews using the GPT-4 Large Language Model}
\author[1]{David Wilkins}
\affil[1]{Discipline of General Practice\\The University of Adelaide}
\maketitle

\begin{abstract}
  Scoping reviews, a type of systematic literature review, require intensive human effort to screen large numbers of scholarly sources for their relevance to the review objectives. This manuscript introduces GPTscreenR, a package for the R statistical programming language that uses the GPT-4 Large Language Model (LLM) to automatically screen sources. The package makes use of the chain-of-thought technique with the goal of maximising performance on complex screening tasks. In validation against consensus human reviewer decisions, GPTscreenR performed similarly to an alternative zero-shot technique, with a sensitivity of 71\%, specificity of 89\%, and overall accuracy of 84\%. Neither method achieved perfect accuracy nor human levels of intraobserver agreement. GPTscreenR demonstrates the potential for LLMs to support scholarly work and provides a user-friendly software framework that can be integrated into existing review pipelines.
\end{abstract}

\section{Introduction}

A scoping review is a relatively novel type of literature review that aims to map the key concepts and existing activity within an area of research \cite{Arksey.2005}. Like systematic reviews, scoping reviews typically use rigorous, transparent \cite{Pham.2014}, and sometimes pre-registered methods for gathering and synthesising evidence, and increasingly use formal frameworks for both performing and reporting reviews \cite{Peters.2021}. Scoping reviews can inform future systematic reviews or primary research in the same area \cite{Sutton.2019}. However, they differ from systematic reviews in aiming to describe the breadth of coverage of the available literature rather than research findings in depth \cite{Arksey.2005}.

Frameworks for performing a scoping review typically involve defining a research area or question, searching bibliographic databases for potentially relevant published material (`sources'), screening these sources to identify those relevant to the area or question, and systematically extracting and reporting data from the relevant sources \cite{JBI.2015, Arksey.2005}. The screening stage will usually involve initial screening of source titles and abstracts against pre-determined inclusion and exclusion criteria, followed by full text screening, with both steps performed in replicate by at least two human reviewers \cite{Peters.2020, JBI.2020, Pham.2014}. Because database searches can return many hundreds or thousands of potentially relevant sources, these screening steps can require intensive human time and effort. Many software methods have been proposed or used to support or partially automate source screening for scoping reviews, including text mining to prioritise relevant sources for human screening \cite{Shemilt.2014, Howard.2016, Chai.2021}, automated clustering and labelling of sources to support human decision-making \cite{Stansfield.2013}, and `crowdsourcing' screening to untrained workers via online platforms \cite{Mortensen.2017}. A similar but more extensive set of methods have been developed and employed for systematic reviews \cite{Khalil.2022, Gates.2019} for which the process of source screening is broadly comparable.

Since the release of the first Generative Pre-trained Transformer (GPT) Large Language Model (LLM) by OpenAI (San Francisco, California, United States of America) in 2018 \cite{Radford.2018}, transformer-based LLMs and the GPT lineage in particular have seen rapid and widespread adoption for a range of automation tasks. Broadly, these models generate a probabilistically weighted list of `tokens' (parts of text such as letter combinations and punctuation) to continue or complete some input text (a `prompt'), having been trained to do so by practising such predictions on large human-written corpora. When this generative process is iterated, it allows for a range of applications involving analysis and production of text, such as summarising documents, generating fiction in a particular genre or style, or conversing with humans \cite{OpenAI.2023}.

While LLMs are not yet widely used to screen sources for literature reviews, early work suggests they may perform well in this role. Guo et al. \cite{Guo.2023} reported the use of a GPT-lineage model (they do not specify which, though their published code suggests OpenAI's `gpt-3.5-turbo' model) to screen 24,307 titles and abstracts from five systematic reviews and one scoping review, achieving weighted average sensitivity of 76\% and specificity of 91\% when compared to human reviewers. Their approach involves giving the model a brief prompt instructing it to take on the persona of a researcher screening titles and abstracts, followed by a source's title and abstract as well as the inclusion and exclusion criteria. The model is instructed to respond with a decision to include or exclude the source, and the process is iterated across the full set of sources to be screened. Syriani et al. \cite{Syriani.2023} similarly reported the use of `gpt-3.5-turbo' to screen titles and abstracts for a systematic review and achieved sensitivities of above 70\%. They also systematically evaluated prompts given to the LLM to identify a prompt that performed best at the screening task; their chosen prompt, like that of Guo et al., placed the LLM in the role of an academic reviewer.

Both of these approaches made use of a single, fixed text prompt template, which the LLM then completes with additional text representing its response (the decision to include or exclude a source), a method sometimes called `zero-shot prompting'. Recent work has identified a number of methods which can be superior to zero-shot prompting when using LLMs for tasks that require complex or multi-step reasoning. These methods include `chain-of-thought prompting' \cite{Wei.2022}, in which a complex task is broken down into a series of intermediate steps, and the `tree of thoughts' strategy \cite{Yao.2023}, in which multiple parallel chains of thought are generated, compared, and integrated. The LLM is induced to follow these complex reasoning strategies either by being given examples of multi-step reasoning on similar tasks, or by being lead through the process with a series of intermediate prompts.

In this paper, I introduce a package for the R programming language \cite{R.2023} called GPTscreenR that implements a chain-of-thought approach to using GPT-4 for scoping review title and abstract screening, and evaluate its performance by comparison to human reviewers. The purpose of this package is to assist and augment rather than replace human reviewers in performing scoping reviews. This package represents the first LLM-based screening tool designed specifically for scoping reviews. Further, this study provides the first report on the accuracy of LLM-based screening using the most recent iteration of the GPT model lineage, GPT-4, and using the recently developed chain-of-thought approach.

\section{Methods}

\subsection{The GPTscreenR package}

GPTscreenR is an R \cite{R.2023} package released under the MIT open source licence. The source code is available for download from GitHub at \url{https://github.com/wilkox/GPTscreenR}. At the time of writing the most recent package version was 0.0.3, and the results presented in this paper were obtained using this version.

GPTscreenR consists of two main components. The first is a set of internal functions for interfacing with the OpenAI Application Programming Interface (API), which allows for `conversations' with the GPT-4 LLM, as well as functions for representing and manipulating those conversations. These internal functions are designed to be model-agnostic, so that future versions of the package or users with particular needs can use different GPT models. The OpenAI API requires an OpenAI account, and OpenAI charges fees for use of the API. In order to access the API, GPTscreenR requires a secret key to be registered prior to source screening, and instructions for doing so are provided in the package documentation and on loading of the package in R if the key has not been correctly registered.

The second component is a set of user-facing functions to perform source screening with GPT-4. The \texttt{review\_description()} function assists in generating a text description of the review's objectives and inclusion and exclusion criteria, using the Population, Concept, and Context (PCC) framework \cite{Peters.2020, JBI.2020} for defining the review's inclusion and exclusion criteria. The use of this function is optional, and users may instead choose to provide a description of the review and criteria for source selection using any framework or format they see fit.

\begin{figure*}
\centering
  \includegraphics[width=\textwidth]{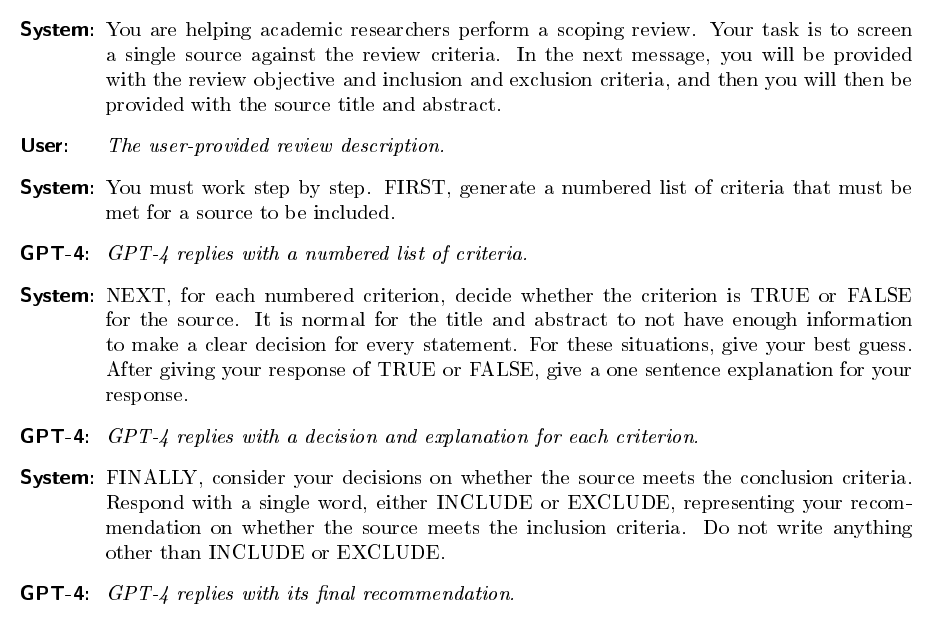}
  \caption{Template for the conversation with GPT-4 mediated by the \texttt{screen\_source()} function. Messages with variable content, including user-provided data as well as GPT-4's responses, are given in italics.}
  \label{fig:conversation_template}
\end{figure*}

The \texttt{screen\_source()} function performs the main task of the package. This function mediates an conversation with GPT-4 in which chain-of-thought prompting \cite{Wei.2022} is used to guide GPT-4 through screening a source title and abstract against the study inclusion criteria. The template for this conversation is given in Figure~\ref{fig:conversation_template}. The OpenAI API defines a conversation as a series of messages, each of which originates from one of three roles: \textit{system}, representing an authoritative voice that can instruct GPT-4 on its task and behaviour; \textit{user}, representing a human user that can interact with GPT-4; and \textit{assistant}, representing the responses generated by GPT-4. In \texttt{screen\_source()}, the \textit{system} role gives GPT-4 general instructions, while the \textit{user} role provides the user-written review description and the source title and abstract.

The specific phrase `work step by step' is derived from the `let's think step by step' prompt phrase which significantly improves LLM performance on multi-step reasoning tasks with a zero-shot prompt \cite{Kojima.2022}, adapted to this chain-of-thought approach.

GPT-4 is instructed to summarise the inclusion criteria for the scoping review, compare the title and abstract against these summarised criteria, and make a final recommendation on whether to include or exclude the source (Figure~\ref{fig:conversation_template}). This approach was chosen after noting that a major source of error when attempting to screen sources with zero-shot, one-shot, or few-shot prompts (i.e.~with a single prompt and no, one, or a few examples) was that GPT-4 would fail to consider important inclusion criteria. As an example, Figure~\ref{fig:zeroshot_fails} presents a conversation with GPT-4 using a zero-shot prompt. The screening task in this example is intentionally adversarial and designed to lead the model towards making an error. In order to correctly recommend exclusion of the source, GPT-4 must notice that the review is looking for research on therapy alpaca interventions, but that the source reports on a therapy camel intervention. The presence of multiple other inclusion criteria which are met by the source, as well as the mention of alpacas in the source abstract, serve as distractors. In this example, GPT-4 incorrectly recommends inclusion. If the conversation is then continued to draw GPT-4's attention to the error, it is able to identify and correct it (Figure~\ref{fig:zeroshot_recognises}), suggesting that the error arises from a failure of GPT-4 to properly consider the relevant inclusion criterion rather than an inability to do so. Using the chain-of-thought approach overcomes this problem without the need for human intervention and correction (Figure~\ref{fig:cot_succeeds}). GPT-4 identifies `The context of the study must involve the use of therapy alpacas as a part of a programme of care in residential aged care facilities.' as an inclusion criterion, and correctly assesses that this is the only inclusion criterion not met. GPT-4 then correctly recommends exclusion of the source.

\setcounter{figure}{0}
\renewcommand\thefigure{2\alph{figure}}

\begin{figure*}
  \centering
  \includegraphics[height=0.85\textheight]{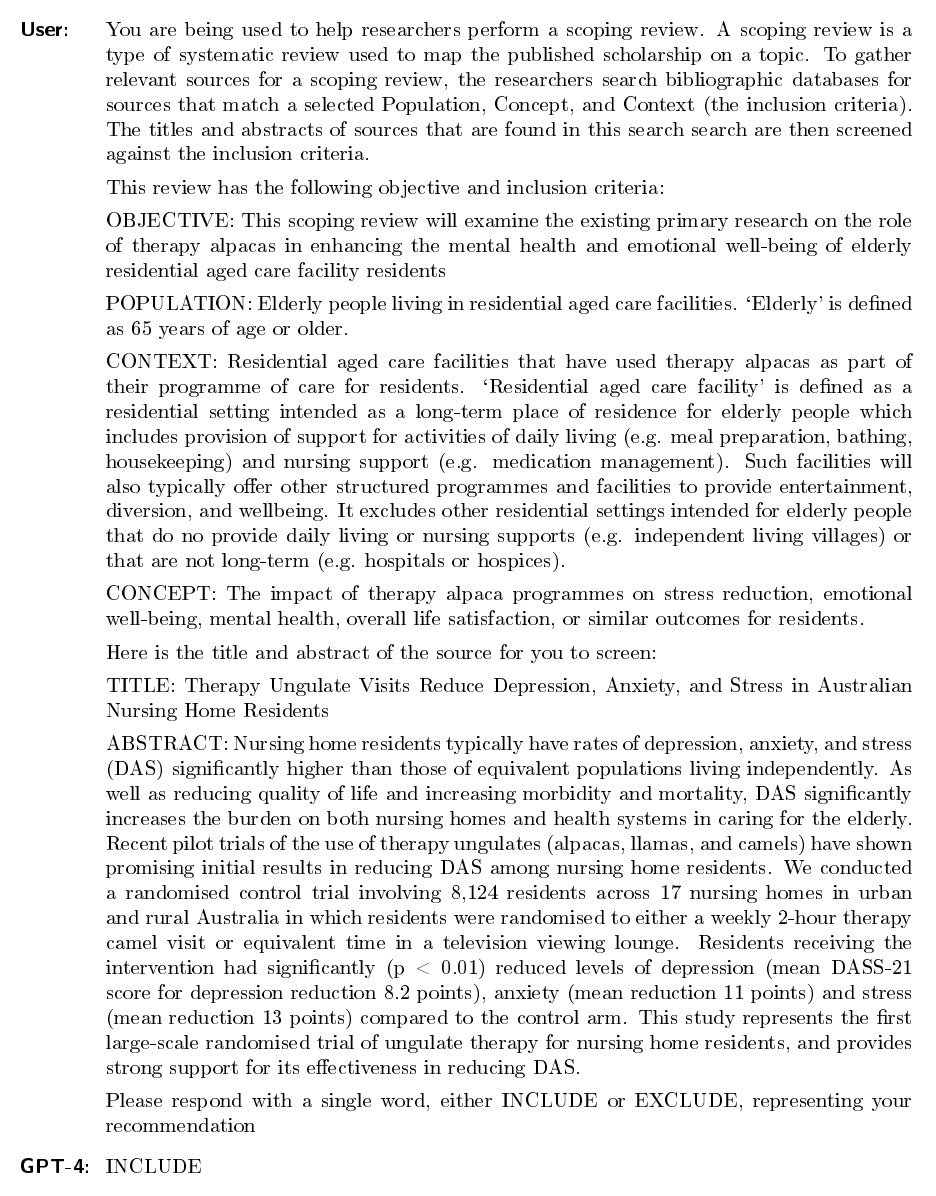}
  \caption{A conversation with GPT-4, using a zero-shot prompt to instruct the model to screen a single source for a hypothetical scoping review. This adversarial task has been intentionally designed to influence GPT-4 towards making an error.}
  \label{fig:zeroshot_fails}
\end{figure*}

\begin{figure*}
  \centering
  \includegraphics[width=\textwidth]{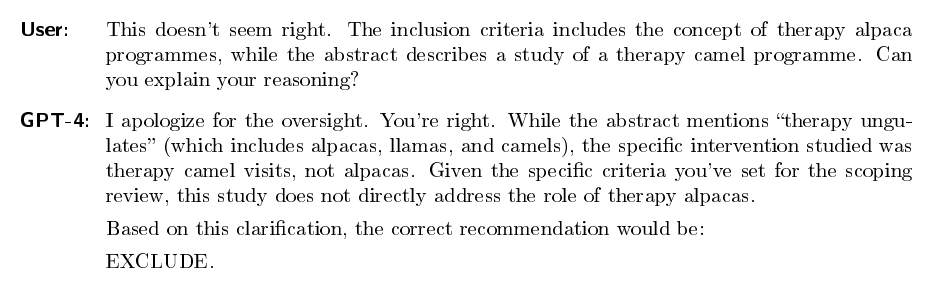}
  \caption{A continuation of the conversation in Figure~\ref{fig:zeroshot_fails}, in which GPT-4 is lead to recognise and correct the error.}
  \label{fig:zeroshot_recognises}
\end{figure*}

\begin{figure*}
\centering
  \includegraphics[width=\textwidth, height=0.85\textheight, keepaspectratio]{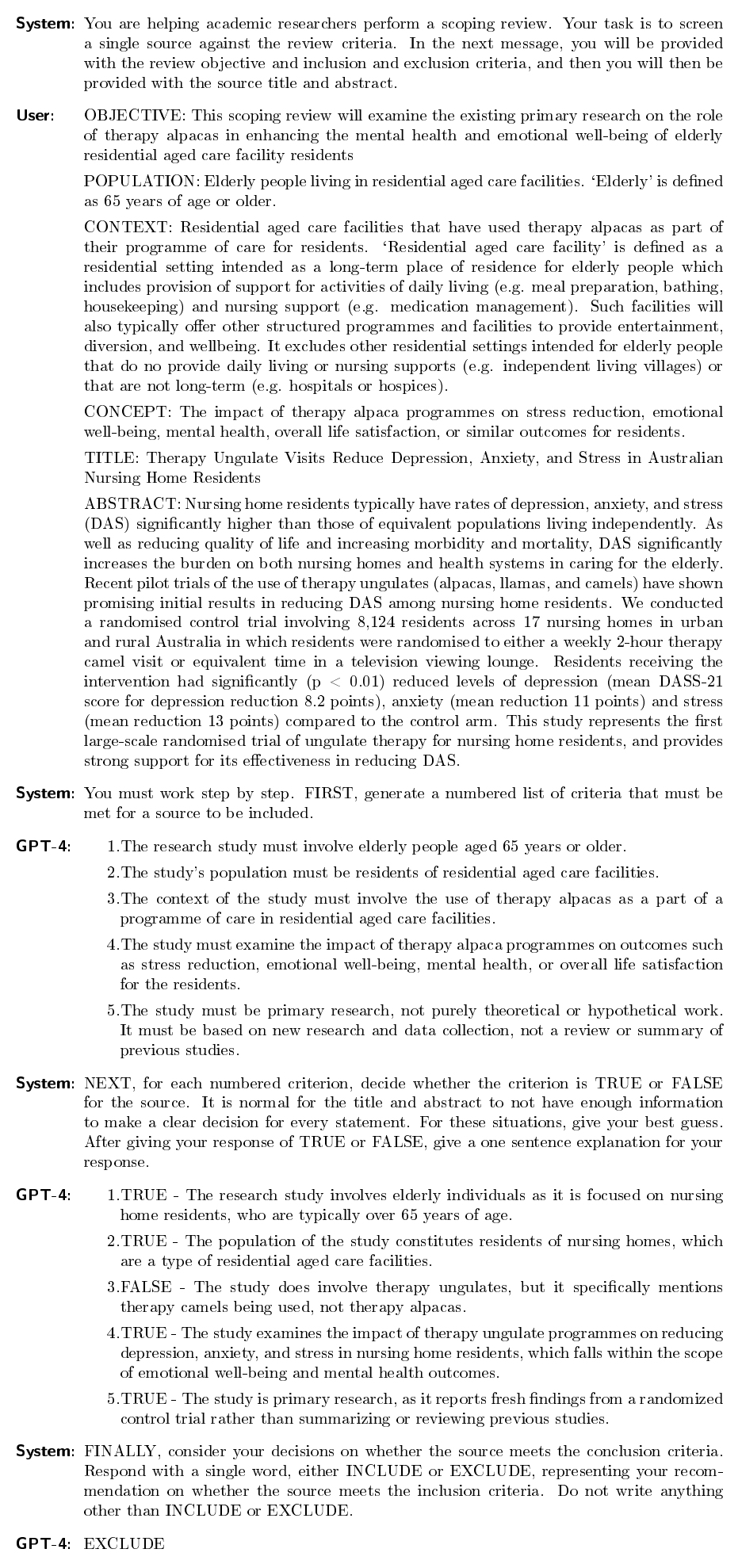}
  \caption{A conversation with GPT-4, instructing it to perform the same adversarial task presented in Figure~\ref{fig:zeroshot_fails}, but using the chain-of-thought approach. This is not a continuation of the conversation presented in Figures~\ref{fig:zeroshot_fails}~and~\ref{fig:zeroshot_recognises} but a new conversation. In this case, GPT-4 correctly recommends excluding the source. The model also correctly identifies that the source meets all of the relevant inclusion criteria except for the requirement that the source report on an alpaca intervention.}
  \label{fig:cot_succeeds}
\end{figure*}

The \texttt{screen\_source()} function returns a list comprising the complete transcript of the conversation with GPT-4 and GPT-4's final recommendation as a logical (Boolean) value. The conversation transcript can be used to interrogate cases where GPT-4 may have returned an incorrect or unexpected result. The package also provides a function \texttt{screen\_sources()}, which applies \texttt{screen\_source()} iteratively across a data frame of sources. \texttt{screen\_sources()} caches screening outcomes to a file as they accumulate, so that screening can be split across multiple sessions and recover from interruptions.

\subsection{Validation}

To validate GPTscreenR's approach, six scoping reviews were identified from the Open Science Framework (OSF; \url{https://osf.io}) where the review inclusion criteria and the results of title and abstract screening were publicly available. A summary of the review characteristics is provided in Table~\ref{tab:validation_reviews}. Random subsets were used where the large number of sources made it prohibitive in time and cost to screen the full set. The total number of sources available for screening and the number used for validation from each review are given in Table~\ref{tab:validation_reviews}.

\begin{table*}[]
  \centering
  \caption{Scoping reviews used to validate the performance of GPTscreenR. `total available' is the number of sources available for screening following attempted retrieval of missing abstracts, correction of malformed data, and removal of sources with missing, irreparable, or duplicate data; it may differ from the number reported by the review authors.}
  \label{tab:validation_reviews}
  \begin{tabularx}{\textwidth}{lXlXp{2cm}}
    \toprule
    Short name & OSF project title & OSF URL & File(s) & Sources screened (total available) \\
    \midrule
    `COVID' & Complementary, Alternative, and Integrative Medicine-Specific COVID-19 Misinformation on Social Media: A Scoping Review & \url{https://osf.io/ytz5e} & `TITLE ABSTRACT FULL TEXT SCREENING DATA\_CAIM COVID-19 SM Misinfo\_Mar2223.xlsx` & 324 (458) \\
    `melanoma' & Scoping Review – Melanoma & \url{https://osf.io/knje4} & `Melanoma Data.xlsx` & 72 (128) \\
    `smartphones' & Smartphone Addiction Scoping Review & \url{https://osf.io/f9huw} & `Scop\_rev\_full.csv` & 256 (5,376) \\
    `solastalgia' & Australian Solastalgia Scoping Review & \url{https://osf.io/qxe3n} & `Solastalgia re-screening and extraction spreadsheet Dec 2022.xlsx` & 150 (150) \\
    `suicide' & Social Norms and Suicidality - Scoping Review & \url{https://osf.io/btpzc} & `June 2022 Screenings Extraction TOP UP.xlsx`, `Nov20 Stage 2 Removal of duplicates and screenings.xlsx` & 100 (2,094) \\
    `teachers' & Teachers' soft skills: a Scoping Review & \url{https://osf.io/n9rkd} & `Teachers´ Soft Skills a SR - Screening NM \& OJ(1).xlsx` & 245 (355) \\
    \bottomrule
  \end{tabularx}
\end{table*}

Some of the reviews did not include source full abstracts in the publicly available files, and when these abstracts could not be obtained from external databases these sources were excluded from validation. There were also many cases where missing, malformed, or duplicate data required either manual correction or exclusion of sources. The scoping review data, code used to prepare this data for validation, and code for calculating summary statistics are available in a reproducible form in the package repository on GitHub (\url{https://github.com/wilkox/GPTscreenR/tree/master/validation}).

The consensus human reviewer decision at the title and abstract screening level was used as the gold standard outcome. Accuracy, sensitivity and specificity were calculated by comparing GPT-4's recommendation against this gold standard. Three of the scoping reviews (\texttt{COVID}, \texttt{solastalgia}, and \texttt{teachers}) included individual human reviewer decisions in addition to the final decision in their publicly available datasets, and these were used to calculate human intraobserver agreement (Cohen's kappa) using the R function \texttt{cohen.kappa()} from the \texttt{psych} package \cite{Revelle.2023}. This was compared to human/GPT-4 agreement across all screened sources, calculated with the same method.

\subsection{Comparison to zero-shot method}

In order to directly compare the chain-of-thoughts approach to a zero-shot approach (i.e.~a conversation consisting of a single prompt with no examples, followed by GPT-4's response), the validation screening task was repeated using the prompt designed by Guo et al. \cite{Guo.2023}, substituting the permitted responses \texttt{INCLUDE} and \texttt{EXCLUDE} for \texttt{included} and \texttt{excluded} respectively in order to maintain compatibility with GPTscreenR's parsing of the response (Figure~\ref{fig:zeroshot_prompt}). In cases where a random subset of sources had been used for validation of GPTscreenR, the same subset was used. The code used to prepare this data and calculate summary statistics was otherwise identical to that used for the chain-of-thoughts method validation and is available in the `zeroshot' branch of the package repository on GitHub (\url{https://github.com/wilkox/GPTscreenR/tree/zeroshot}).

\setcounter{figure}{2}
\renewcommand\thefigure{\arabic{figure}}

\begin{figure*}
\centering
\includegraphics{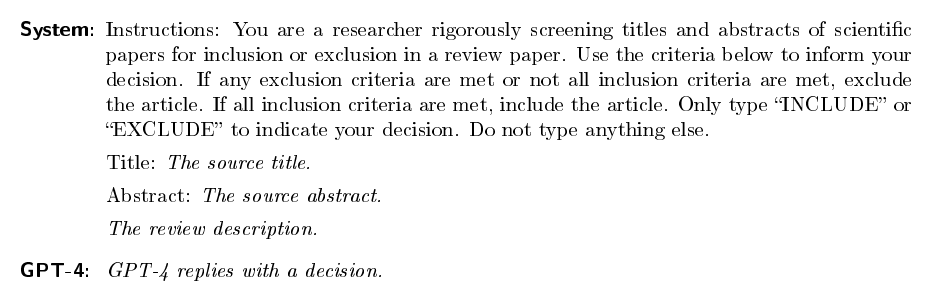}
  \caption{Zero-shot prompt used as a comparator for the chain-of-thoughts approach, derived from the approach of Guo et al. \cite{Guo.2023}. Variable content, including user-provided data as well as GPT-4's response, is given in italics.}
  \label{fig:zeroshot_prompt}
\end{figure*}

\section{Results}

\begin{figure}
\centering
  \includegraphics[width=\columnwidth]{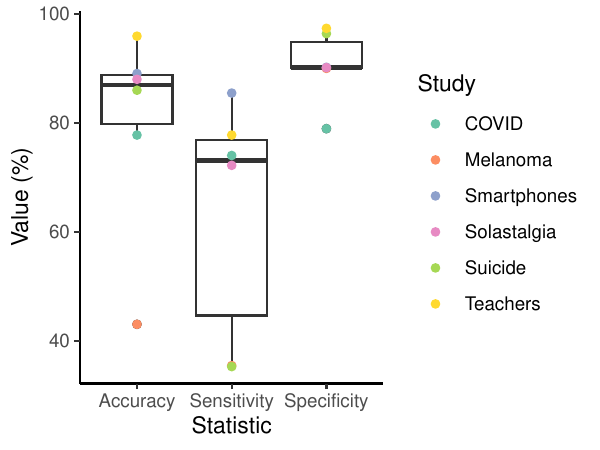}
\caption{Accuracy, sensitivity, and specificity of GPTscreenR compared to the gold standard of the consensus human reviewer decision. The box-and-whisker plot represents for each statistic the median (central bar), 25th and 75th percentiles (upper and lower hinges of the box), and largest value \(\pm\) 1.5 \(\times\) the interquartile range (whiskers). Points given the values for the individual scoping reviews used for validation.}
  \label{fig:statistics}
\end{figure}

1,147 sources were screened from the six scoping reviews. GPTscreenR achieved an overall accuracy of 84\% compared to the gold standard of the consensus human reviewer decision, with weighted average sensitivity of 71\% and weighted average specificity of 89\% (Figure~\ref{fig:statistics}). For the three reviews that provided individual reviewer decisions, the weighted average Cohen's kappa was 0.67, while the weighted average Cohen's kappa between final human and GPT-4 decisions was 0.52.

The zero-shot method achieved an overall accuracy of 83\%, with weighted average sensitivity of 72\% and weighted average specificity of 87\%. The weighted average Cohen's kappa between human and GPT-4 decisions was 0.52.

\section{Discussion}

\subsection{Performance of GPTscreenR}

GPTscreenR and its chain-of-thought approach performed comparably to the zero-shot method reported by Guo et al. \cite{Guo.2023}. Also similarly to their report, Cohen's kappa was higher for inter-human than human-GPT-4 agreement, suggesting that despite this generally good performance GPT-based methods still do not match the performance of a human reviewer when the consensus human decision is used as the gold standard.

\subsection{Comparison to the zero-shot method}

The replication of Guo et al.'s zero-shot prompt on the reviews used for validation in this study resulted in similar weighted average sensitivity (78\%) compared to their report (76\%), though somewhat poorer specificity (84\% compared to 91\%). There are a number of possible factors contributing to this difference. Firstly, this study used the GPT-4 model while Guo et al.~likely used the `gpt-3.5-turbo' model, although it might be expected that the more advanced GPT-4 would generally perform better on the same task. Secondly, this study included only scoping reviews, while Guo et al.~examined five systematic reviews and only one scoping review, although the reported sensitivity (100\%) and specificity (94\%) for that scoping review were higher than the weighted average for this study. Thirdly, validation with both the chain-of-thoughts and zero-shot methods was dependent on the availability of source and screening data from the included reviews, and it was noted that most of the reviews had issues with missing, malformed, missing, and duplicate data, including lacking the text of some or all source abstracts and sufficient information to retrieve them from public databases. While effort was made to rectify these issues, many sources from the included reviews could not be used for validation, and the differing quality of the datasets used in the two studies may have affected the performance of the zero-shot method. Finally, the different reviews used for validation in the two studies likely posed different levels of difficulty the zero-shot approach. If LLM-based approaches to source screening become more widely adopted, the performance of different methods can be more accurately determined across a growing sample of reviews.

Compared against the zero-shot method, the chain-of-thought method achieved poorer sensitivity but higher specificity. This result is consistent with rationale for selecting the chain-of-thought method, which was to reduce type I errors (false positives) of the type demonstrated in Figure~\ref{fig:zeroshot_fails}. However, this came at the cost of higher type II error. It is not possible to retrospectively examine the reasoning process that lead GPT-4 to make an incorrect recommendation such as a false negative, as asking it to explain its reasoning post-hoc will result in a confabulated statistically likely response; as they have no direct access to their internal processes, LLMs are unable to meaningfully introspect. However, unlike the zero-shot method, the chain-of-thought method as implemented by GPTscreenR does allow reviewers to view and assess the reasons for GPT-4's final decision, by reviewing the transcript of the conversation with GPT-4 and in particular its assessment of the source against its summarised list of criteria. This may be useful for resolving disagreements between human reviewers and GPT-4, or disagreements among human reviewers.

\subsection{Limitations of this approach}

Compared to the prompts used by Guo et al. \cite{Guo.2023} and Syriani et al. \cite{Syriani.2023}, both the prompt used for this approach and the response generated by GPT-4 have substantially higher token counts. This results in both a longer time to screen a single source (typically 20--30 seconds) and a higher dollar cost, as the OpenAI API currently charges on a per-token basis. This may make GPTscreenR less appealing to some reviewers, particularly when screening a large number of sources.

Validation of GPTscreenR was limited by a small number of scoping reviews with publicly available, high-quality data on human title and abstract screening decisions. Because a subset of these reviews were used for both testing and validation, there is a risk of over-fitting of the prompt to these particular reviews. The public release of GPTscreenR encourages users to contribute data from their own scoping reviews to support more accurately measuring GPTscreenR's real-world performance, and to guide further refinement of the approach.

\section{Conclusions}

The use of Large Language Models to screen sources for scoping reviews is a promising technology to reduce the human time and effort required to perform these reviews. The GPTscreenR package performs comparably to a zero-shot based approach to source screening. However, as with the zero-shot approach, agreement between its decisions and the human reviewer consensus fell short of inter-human agreement, suggesting that LLM-based screening is not yet as reliable as a human reviewer. Future work may help to further quantify and refine GPTscreenR's performance, and to expand this approach to other types of scholarly review and to other stages of the review process.

\section{References}

\printbibliography

\end{document}